\documentclass[10pt,twocolumn,letterpaper]{article}
\usepackage{graphicx}
\usepackage{subcaption}
\usepackage[pagenumbers]{cvpr} 

\usepackage{amsmath}

\usepackage{listings}
\usepackage{booktabs,makecell, multirow, tabularx}
\usepackage{subcaption}

\definecolor{cvprblue}{rgb}{0.21,0.49,0.74}
\usepackage[pagebackref,breaklinks,colorlinks,allcolors=cvprblue]{hyperref}
\def\paperID{****} 
\def\confName{CVPR}
\def\confYear{2025}

\title{Navigating Image Restoration with VAR’s Distribution Alignment Prior}

\author{Siyang Wang, Feng Zhao\thanks{Corresponding author.} \\
University of Science and Technology of China\\
{\tt\small siyangw@mail.ustc.edu.cn, fzhao956@ustc.edu.cn}
}

\begin{document}
\maketitle

\begin{abstract} 
Generative models trained on extensive high-quality datasets effectively capture the structural and statistical properties of clean images, rendering them powerful priors for transforming degraded features into clean ones in image restoration. VAR, a novel image generative paradigm, surpasses diffusion models in generation quality by applying a next-scale prediction approach. It progressively captures both global structures and fine-grained details through the autoregressive process, consistent with the multi-scale restoration principle widely acknowledged in the restoration community. Furthermore, we observe that during the image reconstruction process utilizing VAR, scale predictions automatically modulate the input, facilitating the alignment of representations at subsequent scales with the distribution of clean images. To harness VAR's adaptive distribution alignment capability in image restoration tasks, we formulate the multi-scale latent representations within VAR as the restoration prior, thus advancing our delicately designed VarFormer framework. The strategic application of these priors enables our VarFormer to achieve remarkable generalization on unseen tasks while also reducing training computational costs. Extensive experiments underscores that our VarFormer outperforms existing multi-task image restoration methods across various restoration tasks. The code is available at \url{ https://github.com/siywang541/Varformer}.

\end{abstract}

\section{Introduction}
\label{sec:intro}

\begin{figure}[!ht]
\centering 
{\includegraphics[width=0.85\linewidth]{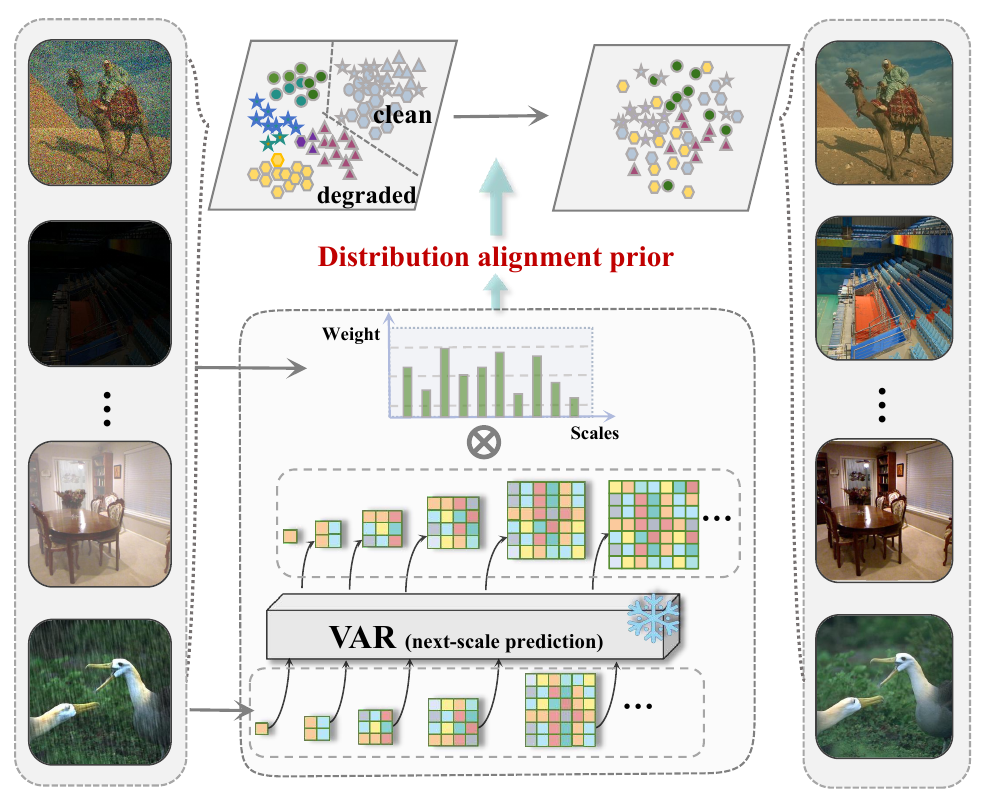}}
\caption{Motivation of VarFormer. (1) As the autoregressive scale evolves, VAR's multi-scale representations shift focus from 
capturing global patterns at lower scales to highlighting fine-grained details at higher scales. 
(2) VAR's scale predictions adaptively modulate the input to align with the distribution of clean images. 
Utilizing VAR's alignment prior on varied scale features related to degradation types allows us to eliminate associated degradations. 
} 
\label{fig:m_comp}
\end{figure}

Image restoration (IR) aims to reconstruct a high-quality (HQ) image from its degraded low-quality (LQ) counterpart, making it widely applicable in various real-world scenarios, including photo processing, autonomous driving, surveillance, segmentation and detection ~\cite{feng2018challenges,urur,chen2024sam2}. 
Recent advances in deep learning have led to powerful IR approaches that excel at addressing specific types of degradation, such as denoising~\cite{wang2023lg,zhang2017beyond,zhang2022practical}, deblurring~\cite{quan2023neumann,ruan2022learning,tsai2022stripformer}, and low-light enhancement~\cite{guo2020zero,wu2023learning,xu2023low}, etc. 
However, these task-specific models struggle with varied and unpredictable degradations in real-world scenarios. This highlights the need for a generalist approach capable of addressing multiple degradation types.
\begin{figure*}[ht]
\centering 
\includegraphics[width=0.9\linewidth]{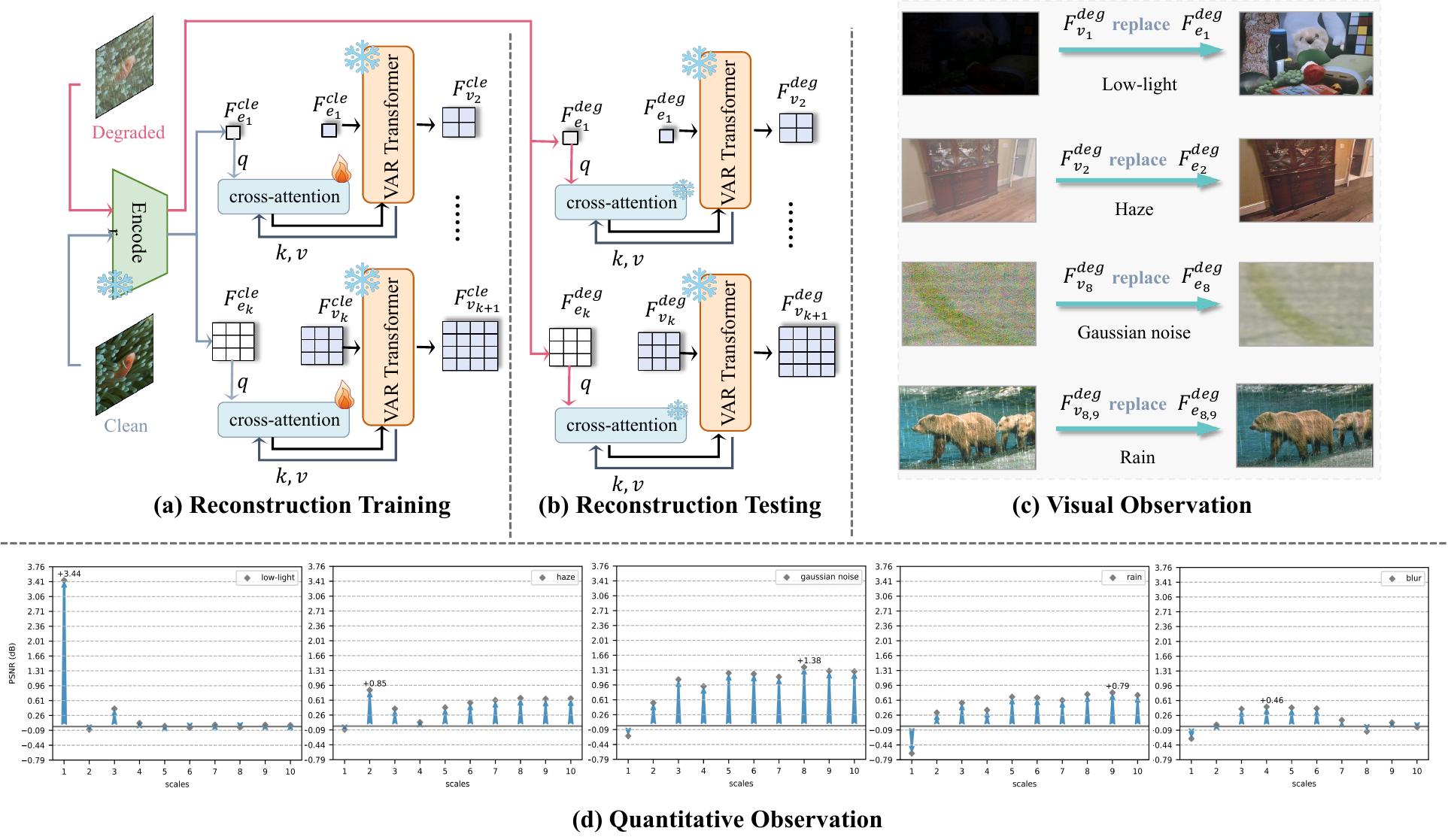}
\caption{Illustration of our investigation about the multi-scale distribution alignment priors within VAR. 
(a) Reconstruction Training: To ensure the coherence of teacher-forcing-based autoregressive predictions, we employ cross-attention to inject multi-scale embeddings ${F}_{e}$ obtained from the VQVAE encoder into the scale autoregression Transformer in VAR~\cite{tian2024visual} for image reconstruction pre-training. 
During this process, we freeze the VAR and train only the cross-attention mechanism using clean images. 
(b) Reconstruction Testing: We feed degraded images into the trained VAR to obtain outputs from the VQVAE encoder, denoted as ${F}_{e}^{deg}$, and the multi-scale predictions from the VAR Transformer, denoted as ${F}_{v}^{deg}$. 
(c-d) Observation: By partially replacing ${F}_{e}^{deg}$ with ${F}_{v}^{deg}$ and mapping the modified multi-scale features back to the pixel space through the decoder, the disappearance of various degradations occurs when replacing different scale features, demonstrating a transition from capturing global information at lower scales to focusing on fine-grained details at higher scales. 
} 
\label{fig2}
\end{figure*}

Recently, several pioneers have sought to develop a universal image restoration model, making considerable strides in the field.
MPRNet~\cite{Zamir2021MultiStagePI} designs a multi-stage architecture that progressively learns image restoration functions for various known degradations.  
AirNet~\cite{Li2022AllInOneIR} distinguishes various image degradation in latent space through a contrastive-based degraded encoder. 
IDR~\cite{zhang2023ingredient} adopts an ingredients-oriented paradigm to investigate the correlation among various restoration tasks. 
However, given the infinite possible solutions for each degraded image, relying exclusively on the degraded image as the sole feature source while neglecting the prior distribution of clean images represents a major limitation for these models in terms of structural reconstruction and realistic texture restoration.

To alleviate these limitations, some studies have focused on the impact of generative priors within the image restoration (IR) community. 
These investigations are grounded in the characteristic that generative models inherently learn high-quality image reconstruction, enabling them to maintain robust high-quality priors that facilitate the transition of degraded features to the clean domain. 
These approaches perform IR tasks using the generative prior~\cite{kawar2022denoising,lin2023diffbir,wang2022zero} from the GAN or diffusion models, achieving superior performance and demonstrating the effectiveness of generative prior for IR tasks.
Recently, a novel generative paradigm, VAR, stands out in the image generation community, recognized for its efficiency and generation capabilities that surpass those of Stable Diffusion~\cite{tian2024visual} .
VAR generates images with progressively increasing resolutions by utilizing marker mapping for next-scale predictions, resolving the conflict between the inherent bidirectional and two-dimensional structural correlations in image patch labeling and the unidirectional nature of the autoregressive model. 

Motivated by the powerful generative capabilities of VAR and the alignment between its endogenous scale representation and the widely recognized multi-scale restoration principle, we aim to explore the rich prior knowledge embedded in the pre-trained VAR across the scale dimension for restoration tasks.
Our investigation reveals the adaptability of the autoregressive scale representation, as the VAR feature representation shifts from capturing wide-ranging global patterns at lower scales to highlighting intricate details at higher scales. 
As described in Fig.~\ref{fig2} (a)-(c), we apply pre-trained VAR to reconstruct the paired clean and degraded images, respectively. 
When we replace the partially scale features of the degraded images obtained from the VQVAE encoder with the autoregressive generated scale features, the degradation of the images gradually diminishes.

For a boarder investigation, we further perform a statistical analysis of the prior knowledge related to high-quality images, utilizing 100 paired images for each type of degradation. 
As illustrated in Fig.~\ref{fig:tsne}, VAR's next-scale prediction projects the multi-scale latent representations of both degraded and clean images into a common space, thereby reducing the distribution gap between various degradations and clean images, demonstrating the model's adaptive distribution alignment capability.

Building on the above observation, we develop a novel framework, named  VarFormer, which integrates multi-scale distribution alignment priors from VAR to restore various degradation types within a single model.
Specifically, to adaptively incorporate the valuable VAR scale priors associated with specific degradation types, we design the Degradation-Aware Enhancement (DAE) module that distinguishes different degradation types and filters out unnecessary interference, thereby providing effective guidance for the restoration process.
Then, to alleviate structural warping and textural distortion resulting from the fusion of high-quality priors and low-quality degraded features, we present the Adaptive Feature Transformation (AFT) module. 
Furthermore, we employs an adaptive mixing strategy to effectively incorporate low-level features from the encoder, enhancing image details while mitigating potential information loss incurred by downsampling operations.
By integrating these designs, our VarFormer not only captures a high-quality feature distribution that enables exceptional generalization on unseen tasks but also accelerates convergence, thus reducing training computational costs.

Our contributions are summarized as follows:

\begin{itemize}
    \item We investigate the multi-scale representations of VAR and reveal its endogenous multi-scale distribution alignment priors, which transition from capturing global color information to focusing on fine-grained details, adaptively aligning the input images with clean images scale by scale as the autoregressive scale evolves.
    \item We propose the VarFormer framework integrated with multi-scale priors for multiple degradation restoration, embracing generalization capability on unseen tasks. To the best of our knowledge, this is the first attempt to explore generative priors from VAR for image restoration. 
    \item Extensive experiments on six image restoration tasks demonstrate the efficiency and effectiveness of VarFormer, including deraining, deblurring, dehazing, low-light image enhancement, Gaussian and real denoising. 
\end{itemize}

\section{Related Work}
\label{sec:Relate}
\subsection{Image Restoration}
The purpose of image restoration is to reconstruct high quality natural images from the degraded images (e.g. noise, blur, rain drops) .
Early methods typically focuses on incorporating various natural image priors along with hand-crafted features for specific degradation removal tasks~\cite{Babacan2009VariationalBB,he2010single}. Recently, deep learning based methods have made compelling progress on various image restoration tasks. For instance, 
DGUNet~\cite{mou2022deep} is proposed based on Proximal Gradient Descent (PGD) algorithm for a gradient estimation strategy without loss of interpretability~\cite{mou2022deep}.
IDR~\cite{zhang2023ingredient} employs an ingredients-oriented paradigm to investigate the correlation among various restoration tasks. TransWeather~\cite{valanarasu2022transweather} designs a transformer-based network with learnable weather type queries to tackle various weather degradation. 

\subsection{Generative Priors in Image Restoration}
The generative models trained on clean data excel at capturing the inherent structures of the image, enabling the generation of images that follow natural image distribution, which helps boost the performance in many low-level vision tasks. Generative Priors of pretrained GANs~\cite{brock2018large,karras2017progressive,karras2019style,karras2020analyzing} is previously exploited by GAN inversion~\cite{abdal2019image2stylegan,gu2020image,gu2020image} , whose primary aim is to find the closest latent codes given an input image. Beyond GANs, Diffusion models have also been effectively used as generative priors in IR ~\cite{kawar2022denoising,lin2023diffbir,wang2024exploiting,wang2022zero}, pushing the frontiers of advanced IR. Our work primarily focuses on generative priors derived from VAR~\cite{tian2024visual}, which has never been explored before.

\section{Method}
\label{sec:method}

In this section, we provide a detailed introduction to our method. We first investigate the properties of VAR at different scales in Sec.~\ref{sec:3.1}. And then, we provide a detailed introduction to the structural design of VarFormer in Sec.~\ref{sec:3.2}.

\subsection{Generative Priors in VAR for IR}
\label{sec:3.1}
Different from traditional autoregressive methods, VAR introduces a novel visual autoregressive modeling paradigm, shifting from ``next-token prediction" to ``next-scale prediction". This paradigm can solve mathematical inconsistencies and structural degradation, which is more optimal for generating highly-structured images. In VAR, each unit predicts an entire token map at a different scale. Starting with a 1 × 1 token map ${r}_{1}$, VAR predicts a sequence of multi-scale token maps $({r}_{1}, {r}_{2},\dots,{r}_{K})$, increasing in resolution. The generation process is expressed as:
\begin{equation}
p({r}_{1}, {r}_{2},\dots,{r}_{K}) = \displaystyle\prod_{k=1}^{K}p( {r}_{k}| {r}_{1}, {r}_{2},\dots,{r}_{k-1}),
\end{equation}

where ${r}_{k} \in {[V]}^{{h}_{k} \times {w}_{k}}$ represents the token map at scale $k$, with dimensions ${h}_{k}$ and ${w}_{k}$, conditioned on previous maps $({r}_{1}, {r}_{2},\dots,{r}_{k-1})$. Each token in ${r}_{k}$ is an index from the VQVAE codebook $V$, which is trained through multi-scale quantization and shared across scales.

\begin{figure}[ht]
\centering 
{\includegraphics[width=0.85\linewidth]{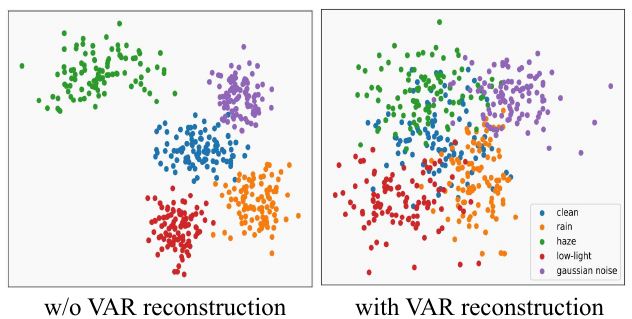}}
\caption{
The t-SNE diagrams demonstrate that VAR's next-scale prediction can reduce the gap between degraded and clean images, effectively aligning their distributions.
} 
\label{fig:tsne}
\end{figure}

To explore the information that VAR focuses on at different scales, we need it to model specific images. First, we obtain GT index sequence ${F}_{e}^{cle} = {[{F}_{{e}_{i}}^{cle}]}_{i=1}^{N}$ at various scales and provide them to VAR. To maintain the continuity and controllability of the VAR modeling process, we inject ${F}_{e}^{cle}$ into the transformer using cross-attention. Finally, we decode the predicted sequence ${F}_{v}^{cle}$ back to the pixel space. The pipeline is shown in Fig.~\ref{fig2}. Please note that the reconstruction training process is conducted only on clean images, with all modules except the cross-attention module being frozen.
Following that, we replace the clean images with degraded images to obtain the GT index sequence ${F}_{e}^{deg} = {[{F}_{{e}_{i}}^{deg}]}_{i=1}^{K}$ and the sequence ${F}_{v}^{deg}$ predicted by the Transformer. To explore the information that VAR focuses on at various scales, we replace the corresponding ${F}_{e}^{deg}$ sequence with the specific scale sequence from ${F}_{v}^{deg}$,Then, we decode the new sequence back into the image space. The process can be mathematically formulated as:
\begin{equation}
 {I}_{rec} = D(\displaystyle\sum_{i\notin C}^{K}{F}_{{e}_{i}}^{deg} + \displaystyle\sum_{i\in C}^{K}{F}_{{v}_{i}}^{deg}),
\end{equation}
where $D$ denotes the VAR Decoder, $C$ refers to the set of indices that need to be replaced. In Fig.~\ref{fig2}, it can be observed that VAR features at different scales naturally restore different types of typical degradation. 
Specifically, substituting lower scales can alleviate global degradations (\emph{e.g.}, low light and haze), while higher scales address local degradations (\emph{e.g.}, noise and rain). Furthermore, as shown in Fig.~\ref{fig:tsne}, by comparing the t-SNE diagrams of the latent representations ${F}{e}$ from the VQVAE encoder and ${F}_{v}$ from the scale autoregression Transformer on various degraded images, the capability of VAR in effectively aligning distributions is demonstrated.
Therefore, the reasonable combination of VAR features at different scales as priors can be beneficial to the image restoration process.

\subsection{Architecture of the VarFormer}
\label{sec:3.2}
\begin{figure*}[ht]
\centering 
\includegraphics[width=0.85\linewidth]{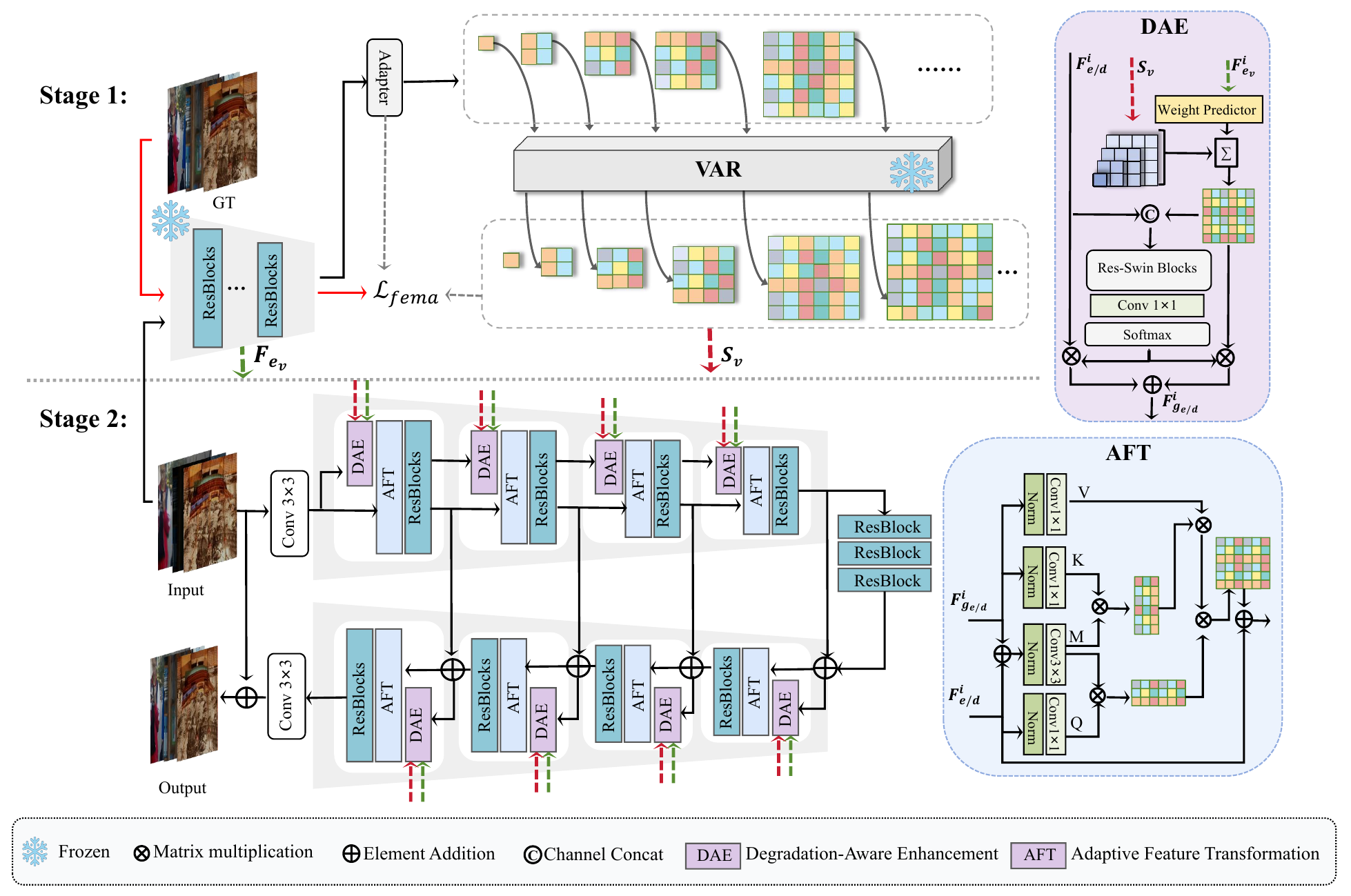}
\caption{The framework of our VarFormer includes two training stages. 
{\bf Stage 1:} To preserve the inherent knowledge of VAR and further enhance its adaptive distribution alignment capability, we freeze the VAR and integrate an Adapter to deliberately reduce the distance between the multi-scale latent representations of clean and degraded images, thereby obtaining multi-scale distribution alignment embedding ${S}_{v}$. 
{\bf Stage 2:} To adaptively extract valuable VAR scale priors for input-specific degradation type, the Degradation-Aware Enhancement (DAE) module is designed to distinguish different degradation types and integrate relevant priors, thus providing effective scale-aware alignment prior for the restoration process. 
Furthermore, the Adaptive Feature Transformation (AFT) module integrates the VAR scale priors into the image restoration network to guide the elimination of degradation.
} \label{fig3}
\end{figure*}

Based on the observations from Sec.~\ref{sec:3.1}, we propose VarFormer, which leverages the multi-scale distribution alignment priors in VAR to facilitate image restoration. As shown in Fig.~\ref{fig3}, the VarFormer training process is divided into two stages. The first stage is dedicated to extracting the priors from VAR, while the second stage utilizes these priors to guide the restoration process. {\bf In the first stage}, the model is enhanced with an Adapter module on top of the VAR that has been finetuned for reconstruction in Sec.~\ref{sec:3.1}. {\bf In the second stage}, the model incorporates Degradation-Aware Enhancement (DAE) modules and Adaptive Feature Transformation (AFT) modules, which integrate the multi-scale distribution alignment priors extracted in the first phase to accomplish image restoration.

{\bf Adapter for Domain Shift} (Train in Stage 1).
Due to the existence of distributional shift between high quality sources for VAR pre-training and degraded datasets for image restoration, direct reuse of the features from the Encoder of VAR may be suboptimal. However, fine-tuning disrupts the knowledge in VAR. Therefore, inspired by AWRCP~\cite{ye2023adverse}, we chose to freeze the encoder and insert an Adapter containing self-attention blocks after it to retain pre-trained knowledge while narrowing the domain gap.

{\bf Degradation-Aware Enhancement (DAE)} (Train in Stage 2).
To address various degradation tasks by appropriately combining information at different scales, we propose Degradation-Aware Enhancement (DAE). An illustration of DAE is shown in Fig.~\ref{fig3}. 
After reconstruction pre-training on extensive clean images, the encoder of VAR has developed the capability to discern what level of information each network layer should focus on. Consequently, we can use the outputs ${F}_{{e}_{v}}$ from its different layers to determine the weights for scale priors. Specifically, for the $i$-th layer of the encoder or decoder in the second stage, we take ${F}_{{e}_{v}}^{i}$ as the input for the weight predictor within the DAE. After processing through Swin-transformer blocks that mitigate the impact of image content and a projection convolutional layer, we derive the weight prediction set $W={[{w}_{i}]}_{i=1}^{K}$.
\begin{equation}
\widehat{S}_{w}^{i} =\mathcal{M}(\displaystyle\sum_{j=1}^{K}{w}_{j}\cdot{S}_{v}^{j}) ,
\end{equation}
where the $\mathcal{M}(\cdot)$  is a lightweight projection head designed to perform dimensionality transformation on the re-weighted priors, the $\widehat{S}_{w}^{i}\in {R}^{{C}^{i}\times{H}^{i}\times{W}^{i}}$ is the re-weighted priors for $i$-th encoder or decoder layer.

Furthermore, we must consider that not every region in a degraded image may need to undergo the transition from the degraded domain to the clean domain. To adjust the guidance strength of the priors according to the degradation conditions of different regions, inspired by ~\cite{liu2024hcanet}, we facilitate interaction between ${F}_{e/d}^{i} \in {R}^{{C}^{i}\times{H}^{i}\times{W}^{i}}$ and $\widehat{S}_{w}^{i}$ at the channel level, enabling them to better capture long-range dependencies. After processing through a lightweight network, we obtain region-specific fusion weights ${w}_{1}^{g}$ and ${w}_{2}^{g}$ for  ${F}_{e/d}^{i}$ and $\widehat{S}_{w}^{i}$, respectively. Ultimately, the feature ${F}_{{g}_{e/d}}^{i}\in {R}^{{C}^{i}\times{H}^{i}\times{W}^{i}}$ is derived by summing ${F}_{e/d}^{i}$ and $\widehat{S}_{w}^{i}$, each multiplied by their respective weights:
\begin{equation}
{w}_{1}^{g}, {w}_{2}^{g} = Softmax(Conv(RSTBs(Concat({F}_{e/d}^{i},\widehat{S}_{w}^{i})))) \nonumber,
\end{equation}
\begin{equation}
{F}_{{g}_{e/d}}^{i} = {F}_{e/d}^{i} \times {w}_{1}^{g} + \widehat{S}_{w}^{i} \times {w}_{2}^{g} ,
\end{equation}
where $Concat(\cdot)$ refers to the concatenation operation, the $RSTBs(\cdot)$ means a series of Residual Swin-Transformer Blocks, the $Conv(\cdot)$ represents the convolutional layer and the $Softmax(\cdot)$ means the Softmax activation layer.

{\bf Adaptive Feature Transformation (AFT)} (Train in Stage 2).
We employ Adaptive Feature Transformation (AFT) to alleviate structural warping and textural distortion resulting from the fusion of high-quality priors and low-quality degraded features. Unlike standard cross-attention, we introduce a low-dimensional intermediate feature $M$ that aggregates both ${F}_{{g}_{e/d}}^{i}$ and ${F}_{e/d}^{i}$ as the mediator for similarity comparison, as illustrated in Fig.~\ref{fig3}. Specifically, given the intermediate feature ${F}_{e/d}^{i}$ and the degradation-aware enhanced feature ${F}_{{g}_{e/d}}^{i}$, we obtain the bridging feature ${F}_{{m}_{e/d}}^{i}=Concat({F}_{{g}_{e/d}}^{i},{F}_{e/d}^{i})$, where ${F}_{{m}_{e/d}}^{i}\in {R}^{{2C}^{i}\times{H}^{i}\times{W}^{i}}$. Then, ${F}_{e/d}^{i}$ is projected into the query $Q={W}_{q}{F}_{e/d}^{i}$, ${F}_{{m}_{e/d}}^{i}$ is projected into the mediator $M={W}_{m}{F}_{{m}_{e/d}}^{i}$, and ${F}_{{g}_{e/d}}^{i}$ is projected into the key $K={W}_{k}{F}_{{g}_{e/d}}^{i}$ and value $V={W}_{v}{F}_{{g}_{e/d}}^{i}$. Here, ${W}_{m}$, ${W}_{q}$, ${W}_{k}$, ${W}_{v}$ are all implemented using convolutional kernels. Formally, the transformation process is defined by the following equations:

\begin{equation}
{F}_{in}^{i+1}={A}_{q,m}\cdot({A}_{m,k}\cdot V) + {F}_{e/d}^{i}\nonumber,
\end{equation}
\begin{equation}
{A}_{q,m}=Softmax(Q\cdot {M}^{T}/\sqrt{d}),\
\end{equation}
\begin{equation}
{A}_{m,k}=Softmax(M\cdot {K}^{T}/\sqrt{d})\nonumber,
\end{equation}
where $L \ll  H$, ${A}_{q,m} \in {R}^{H\times L} $ and ${A}_{m,k} \in {R}^{L\times H} $ denotes the attention map between the query-mediator pair and mediator-key pair.

In addition, to address the inherent loss of detail and texture due to downsampling operations, we propose an adaptive mix-up skipping to integrate encoder features into the decoder, thereby mitigating information loss:
\begin{equation}
{F}_{d}^{i} = \sigma ({\theta }_{i})\times \widehat{F}_{d}^{i} + (1-\sigma ({\theta }_{i}))\times{F}_{e}^{i},
\end{equation}
where ${\theta }_{i}$ represents a learnable coefficient, $\sigma$ denotes the sigmoid operator, and $\widehat{F}_{d}^{i}\in {R}^{{C}^{i}\times{H}^{i}\times{W}^{i}}$ is the output of residual blocks of the $i$-th layer.

\begin{figure*}[!h]
\centering 
\includegraphics[width=0.97\linewidth]{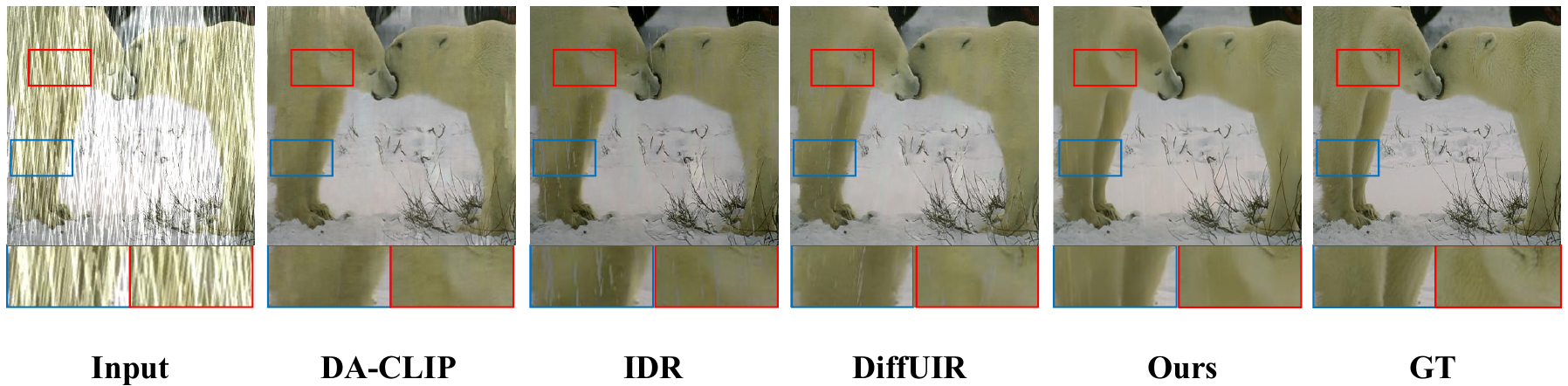}
\caption{Visual comparison with state-of-the-art methods on image deraining task. Please zoom in for details.} 
\label{fig:rain}
\end{figure*}

\begin{figure*}[]
\centering 
\includegraphics[width=0.97\linewidth]{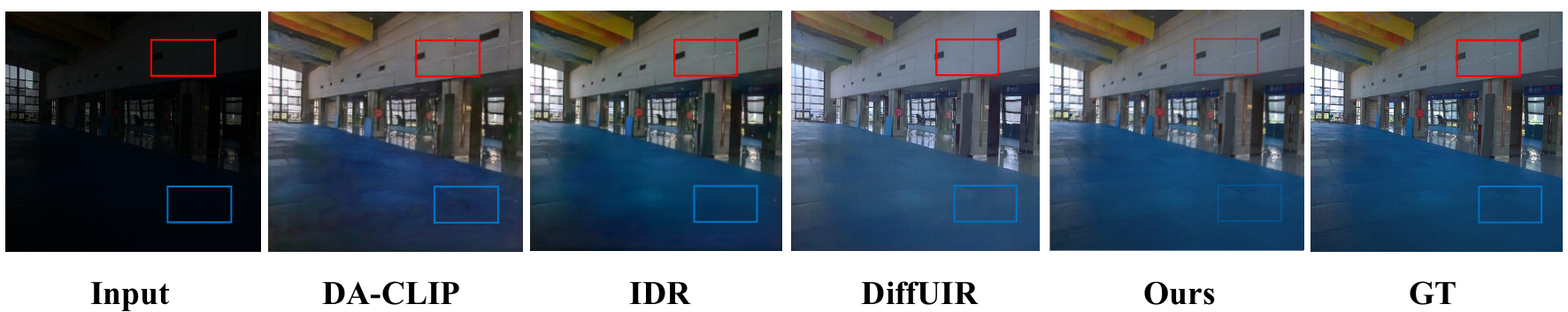}
\caption{Visual comparison with state-of-the-art methods on low-light image enhancement task. Please zoom in for details.} 
\label{fig:blur}
\end{figure*}

\subsection{Training Objectives}

{\bf Feature Matching Loss.} The goal of the first stage is to further enhance the distribution alignment capability of VAR. Consequently, we design the Feature Matching Loss to reduce the feature discrepancies between degraded images and clean images, which can be formulated as follows:

\begin{equation}
{\mathcal{L}}_{fema} = \displaystyle\sum_{i=1}^{K}-{s}_{i}log(\widehat{s}_{i}) + {\left \| 
 {F}_{a}-sg({F}_{{e}_{gt}}^{q}) \right \|}_{2}^{2},
\end{equation}
where ${s}_{i}$ is the  ground-truth of $i$-th scale latent representation, $\widehat{s}_{i}$ is the output of VAR Transformer, and ${F}_{a}$ is the output of Adapter.

\noindent {\bf Reconstruction Loss.} This loss is utilized in the second training stage, with the purpose of ensuring that the restored image possesses a completed structure and impressing visual pleasure. It consists of two components: the PSNR loss and the perceptual loss. It is formulated as follows:
\begin{equation}
{\mathcal{L}}_{rec} = {-PSNR({I}_{gt},{I}_{rec}) + {\left \| \psi ({I}_{gt})-\psi ({I}_{rec})\right \|}_{2}^{2}},
\end{equation}
where $\psi(\cdot)$ indicates  the pre-trained VGG19 network.

\section{Experiments}
\label{sec:exp}

In this section, we first clarify the experimental settings of our method, and then present qualitative and quantitative results compared with the state-of-the-art methods on all-in-one and zero-shot tasks. Moreover, we also conduct extensive ablations to verify the effectiveness of our method.


\begin{table*}[h]
\centering
\caption{Quantitative comparison with task-specific methods and universal methods on image deraining, low-light image enhancement, motion deblurring and image dehazing tasks. $\dagger$ means reimplementing in our datasets with the same settings for fair comparison.
}
\setlength{\tabcolsep}{3mm}{
\begin{tabular}{cccccccccc}
\toprule

\multicolumn{1}{c|}{\multirow{2}{*}{\textbf{Method}}} & \multicolumn{1}{c|}{\multirow{2}{*}{\textbf{Year}}} & \multicolumn{2}{c|}{\textbf{Deraining (5sets)}}    & \multicolumn{2}{c|}{\textbf{Deblurring}}    & \multicolumn{2}{c|}{\textbf{Enhancement}}    & \multicolumn{2}{c}{\textbf{Dehazing}}     \\

\multicolumn{1}{l|}{}                                 & \multicolumn{1}{c|}{}                               & PSNR↑        & \multicolumn{1}{c|}{SSIM↑}        & PSNR↑     & \multicolumn{1}{c|}{SSIM↑}     & PSNR↑      & \multicolumn{1}{c|}{SSIM↑}     & PSNR↑              & \multicolumn{1}{c}{SSIM↑}                \\ \midrule
\multicolumn{10}{l}{\textbf{Task-specific Method}}                                                                                                                                                                                                                                          \\ \midrule
\multicolumn{1}{c|}{SwinIR~\cite{liang2021swinir}}                           & \multicolumn{1}{l|}{2021}                           & -           & \multicolumn{1}{c|}{-}           & 24.52    & \multicolumn{1}{c|}{0.773}    & 17.81     & \multicolumn{1}{c|}{0.723}    & 21.50      & \multicolumn{1}{c}{0.891}                  \\

\multicolumn{1}{c|}{MIRNet~\cite{zamir2020learning}}                        & \multicolumn{1}{l|}{2022}                           & -           & \multicolumn{1}{c|}{-}           & 26.30    & \multicolumn{1}{c|}{0.799}    & 24.74     & \multicolumn{1}{c|}{0.851}    & 24.03      & \multicolumn{1}{c}{0.927}                  \\
\multicolumn{1}{c|}{Restormer~\cite{zamir2022restormer} }                       & \multicolumn{1}{l|}{2022}                           & 33.96       & \multicolumn{1}{c|}{0.935}       & 32.92    & \multicolumn{1}{c|}{0.961}    & 20.41     & \multicolumn{1}{c|}{0.806}    & 30.87         & \multicolumn{1}{c}{0.969}               \\
\multicolumn{1}{c|}{MAXIM~\cite{tu2022maxim}}                            & \multicolumn{1}{l|}{2022}                           & 33.24       & \multicolumn{1}{c|}{0.933}       & 32.86    & \multicolumn{1}{c|}{0.940}    & 23.43     & \multicolumn{1}{c|}{0.863}    & -         & \multicolumn{1}{c}{-}                      \\
\multicolumn{1}{c|}{RDDM~\cite{liu2024residual}}                             & \multicolumn{1}{l|}{2023}                           & 30.65       & \multicolumn{1}{c|}{0.901}       & 28.83    & \multicolumn{1}{c|}{0.846}    & 24.22     & \multicolumn{1}{c|}{0.889}    & 30.76       & \multicolumn{1}{c}{0.943}                  \\ \midrule
\multicolumn{10}{l}{\textbf{Universal Method}}                                                                                                                                                                                                                                              \\ \midrule
\multicolumn{1}{c|}{AirNet~\cite{Li2022AllInOneIR}$\dagger$}                            & \multicolumn{1}{l|}{2022}                           & 25.44       & \multicolumn{1}{c|}{0.743}       & 27.14    & \multicolumn{1}{c|}{0.832}    & 18.49     & \multicolumn{1}{c|}{0.767}    & 25.48      & \multicolumn{1}{c}{0.944}                 \\
\multicolumn{1}{c|}{Painter~\cite{wang2023images} }                          & \multicolumn{1}{l|}{2022}                           & 29.49       & \multicolumn{1}{c|}{0.868}       & -        & \multicolumn{1}{c|}{-}        & 22.40     & \multicolumn{1}{c|}{0.872}    & -       & \multicolumn{1}{c}{-}                         \\
\multicolumn{1}{c|}{IDR~\cite{zhang2023ingredient}$\dagger$}                              & \multicolumn{1}{l|}{2023}                           & 30.87           & \multicolumn{1}{c|}{0.906}           & 27.94    & \multicolumn{1}{c|}{0.848}    & 22.32     & \multicolumn{1}{c|}{0.836}    & 25.33      & \multicolumn{1}{c}{0.945 }                \\
\multicolumn{1}{c|}{DA-CLIP~\cite{luo2023controlling}$\dagger$}                        & \multicolumn{1}{l|}{2023}                           & 28.75       & \multicolumn{1}{c|}{0.844}       & 26.24    & \multicolumn{1}{c|}{0.801}    & 24.27     & \multicolumn{1}{c|}{0.885}    & 31.42       & \multicolumn{1}{c}{0.941}                 \\
\multicolumn{1}{c|}{Prompt-IR~\cite{potlapalli2306promptir}$\dagger$}                        & \multicolumn{1}{l|}{2023}                           & 29.44       & \multicolumn{1}{c|}{0.848}       & 27.31    & \multicolumn{1}{c|}{0.834}    & 22.95     & \multicolumn{1}{c|}{0.844}    & 32.17       & \multicolumn{1}{c}{0.953}                  \\
\multicolumn{1}{c|}{DiffUIR~\cite{zheng2024selective}$\dagger$}                        & \multicolumn{1}{l|}{2024}                           & 31.14       & \multicolumn{1}{c|}{0.907}       & 29.88    & \multicolumn{1}{c|}{0.874}    & 25.02     & \multicolumn{1}{c|}{0.901}    & 32.74       & \multicolumn{1}{c}{0.944}                \\

\rowcolor[HTML]{E2E2E2} \multicolumn{1}{c|}{\textbf{VarFormer}}               & \multicolumn{1}{c|}{-}                              &{\color{red}31.33}             & \multicolumn{1}{c|}{{\color{red}0.913}}            &{\color{red}30.99}          & \multicolumn{1}{c|}{{\color{red}0.956}}         &{\color{red}25.13}           & \multicolumn{1}{c|}{{\color{red}0.917}}         &{\color{red}32.96}                   & \multicolumn{1}{c}{{\color{red}0.956}}                       \\ \bottomrule
\end{tabular}}
\label{tab:aio}
\end{table*}

\begin{table}[h]
\centering
\caption{Quantitative results of Gaussian denoising on BSD68,
Urban100 and Kodak24 datasets in terms of PSNR↑.}
\setlength{\tabcolsep}{0.3mm}
\fontsize{9}{12}\selectfont{
\begin{tabular}{c|ccc|ccc|ccc}
\toprule
\multirow{2}{*}{\textbf{Method}} & \multicolumn{3}{c|}{\textbf{BSD68}} & \multicolumn{3}{c|}{\textbf{Urban100}} & \multicolumn{3}{c}{\textbf{Kodak24}} \\
                                 & $\sigma$=15         & $\sigma$=25         & $\sigma$=50        & $\sigma$=15          & $\sigma$=25          & $\sigma$=50         & $\sigma$=15         & $\sigma$=25         & $\sigma$=50         \\ \midrule
HINet~\cite{chen2021hinet}                            & 33.72      & 31.00      & 27.63     & 33.49       & 30.94       & 27.32      & 34.38      & 31.84      & 28.52      \\
MPRNet~\cite{Zamir2021MultiStagePI}                            & 34.01      & 31.35      & 28.08     & 34.13       & 31.75       & 28.41      & 34.77      & 32.31      & 29.11      \\
MIRV2~\cite{zamir2020learning}                         & 33.66      & 30.97      & 27.66     & 33.30       & 30.75       & 27.22      & 34.29      & 31.81      & 28.55      \\
SwinIR~\cite{liang2021swinir}                           & 33.31      & 30.59      & 27.13     & 32.79       & 30.18       & 26.52      & 33.89      & 31.32      & 27.93      \\
Restormer~\cite{zamir2022restormer}                        & 34.03      & 31.49      & 28.11     & 33.72       & 31.26       & 28.03      & 34.78      & 32.37      & 29.08      \\ \midrule
AirNet~\cite{Li2022AllInOneIR}$\dagger$                          & 33.49      & 30.91      & 27.66     & 33.16       & 30.83       & 27.45      & 34.14      & 31.74      & 28.59      \\
IDR~\cite{zhang2023ingredient}$\dagger$                                & 34.05      & 31.67      & 28.05     & 32.92       & 31.29       & 28.45      & 34.53      & 32.22      & 28.93      \\
DA-CLIP~\cite{luo2023controlling}$\dagger$   & 30.16      & 28.89      & 27.04     & 33.14       & 30.99       & 27.61     & 33.86     & 32.30      & 28.84      \\
Prompt-IR~\cite{potlapalli2306promptir}$\dagger$ & 32.17      & 29.89      & 28.03     & 33.04       & 31.89       & 27.72     & 33.78     & 32.21      & 28.64      \\
DiffUIR~\cite{zheng2024selective}$\dagger$       & 33.86      & 30.88      & 26.63     & 32.19       & 29.65       & 25.87      & 33.24      & 30.70      & 27.19      \\
\rowcolor[HTML]{E2E2E2} \textbf{VarFormer}                        & {\color{red}34.11}      & {\color{red}31.85}      & {\color{red}28.23}     & {\color{red}33.13}       & {\color{red}31.94}       & {\color{red}28.96}      & {\color{red}34.73}      & {\color{red}32.40}      & {\color{red}29.02}            \\ \bottomrule
\end{tabular}
}
\label{tab:syn_noise}
\end{table}


\begin{table}[!h]
\centering
\caption{Quantitative results of Real image denoising on SIDD in terms of PSNR↑ and SSIM↑.}
\setlength{\tabcolsep}{3mm}{
\begin{tabular}{c|cc}
\toprule
\multirow{2}{*}{\textbf{Method}} & \multicolumn{2}{c}{\textbf{SIDD}} \\
                                 & PSNR↑                     & SSIM↑                    \\ \midrule
MPRNet~\cite{Zamir2021MultiStagePI}                            & 39.71                    & 0.958                   \\
Uformer~\cite{Wang2021UformerAG}                          & 39.89                    & 0.961                   \\
Restormer~\cite{zamir2022restormer}                        & 40.02                    & 0.967                   \\
ART~\cite{zhang2022accurate}                              & 39.99                    & 0.964                   
                                 \\ \midrule
AirNet~\cite{Li2022AllInOneIR}$\dagger$                          & 38.32                    & 0.945                   \\
Painter~\cite{wang2023images}                          & 38.88                    & 0.954                   \\
IDR~\cite{zhang2023ingredient}$\dagger$                          & 39.74                    & 0.957                    \\
DA-CLIP~\cite{luo2023controlling}$\dagger$                          & 34.04                    & 0.824                   \\
Prompt-IR~\cite{potlapalli2306promptir}$\dagger$                         & 39.52                    & 0.954                   \\

DiffUIR~\cite{zheng2024selective}$\dagger$                       & 40.11                    & 0.976                   \\

\rowcolor[HTML]{E2E2E2} \textbf{VarFormer}                        &{\color{red}40.13}                          &{\color{red}0.978}                         \\ \bottomrule
\end{tabular}}
\label{tab:real_noise}
\end{table}

\subsection{Experimental Settings }
{\bf Datasets.} (1) All-in-one:  We train a unified model to solve 6 IR tasks, including deraining, dehazing, low-light enhancement, motion deblurring, Gaussian denoising and real image denoising. For deraining, we adopt Rain13K~\cite{Fu2016ClearingTS,Li2018RecurrentSC,Li2016RainSR,Luo2015RemovingRF,Yang2019SingleID} for training, and 5 datasets for testing, including Rain100L~\cite{Yang2016DeepJR}, Rain100H~\cite{Yang2016DeepJR}, Test100~\cite{Zhang2017ImageDU}, Test1200~\cite{Zhang2018DensityAwareSI} and Test2800~\cite{Fu2017RemovingRF}. For dehazing, the RESIDE~\cite{li2018benchmarking} dataset is used as the benchmark. As the real-world fog condition is outdoor, we only train and test on the outdoor part. For low-light enhancement, LOL dataset~\cite{Wei2018DeepRD} is adopted. For motion deblurring, we adopt GoPro~\cite{Nah2016DeepMC} dataset for training and testing.
For Gaussian denoising, BSD400~\cite{martin2001database} and WED~\cite{ma2016waterloo} are used for training and BSD68~\cite{martin2001database}, Urban100~\cite{Huang2015SingleIS}, Kodak24~\cite{franzen1999kodak} for testing.
For real image denoising, We use SIDD~\cite{abdelhamed2018high} datasets as the benchmark for training and testing.
(2) Zero-shot: we utilize TOLED~\cite{zhou2021image} and POLED ~\cite{zhou2021image} for under-display camera (UDC) IR.

{\bf Evaluation Metrics.} 
To evaluate the restoration performance, we adopt Signal to Noise Ratio (PSNR), Structural Similarity (SSIM) and Learned Perceptual Image Patch Similarity (LPIPS)~\cite{zhang2018unreasonable}. 
We emphasize the performance of the universal methods, with the best results shown in {\color{red}red}.

{\bf Implementation Details.} 
We adopt the Adam optimizer~\cite{Kingma2014AdamAM}(${\beta }_{1}=0.9$, ${\beta}_{2}=0.999$) with the initial learning rate 1e-4 gradually reduced to 1e-6 with cosine annealing to train our model. We random crop $256\times256$ patch from original image as network input after data augmentation. As the data size varies greatly from task to task, we set the weight of each task in one batch as 0.3 for dehazing, 0.1 for low-light, 0.2 for deraining, 0.2 for Gaussian denoising,0.1 for real image denoising, and 0.1 for motion deblurring.

\subsection{All-in-One Restoration}
We compare VarFormer with both task-specific and universal methods. To ensure a fair evaluation, we train all-in-one models from scratch using our training strategy. The quantitative results are presented in ~\cref{tab:aio,tab:syn_noise,tab:real_noise}, where it can be observed that VarFormer demonstrates superior performance across all 6 tasks. Notably, in low-light enhancement and dehazing tasks, VarFormer's performance with the all-in-one training strategy even surpasses that of task-specific methods, highlighting its exceptional capabilities. We also provide visual comparisons with other state-of-the-art universal methods, as shown in Fig.~\ref{fig:rain} and Fig.~\ref{fig:blur}, with more results available in the supplementary material. It is evident that, compared to other universal methods, VarFormer yields more steady results in all image restoration tasks.

\subsection{Zero-shot Results}
In Tab.~\ref{tab:zero-shot}, we present the generalization performance of various methods on unseen tasks without fine-tuning. Specifically, high-resolution images captured under under-display camera (UDC) systems often suffer from various types of degradations due to the point spread function and lower light transmission rate, posing challenges for the prediction of restoration models. Encouragingly, our VarFormer achieves state-of-the-art performance.

\begin{table}[!h]
\centering
\caption{Quantitative results of unknown tasks (under-display
camera image restoration) on TOLED and POLED datasets.}
\setlength{\tabcolsep}{0.1mm}
\begin{tabular}{c|ccc|ccc}
\toprule
\multirow{2}{*}{\textbf{Method}} & \multicolumn{3}{c|}{\textbf{TOLED}} & \multicolumn{3}{c}{\textbf{POLED}} \\
                                 & PSNR↑       & SSIM↑       & LPIPS↓     & PSNR↑       & SSIM↑      & LPIPS↓     \\ \midrule
NAFNet~\cite{chen2022simple}                           & 26.89      & 0.774      & 0.346     & 10.83      & 0.416     & 0.794     \\
HINet~\cite{chen2021hinet}                            & 13.84      & 0.559      & 0.448     & 11.52      & 0.436     & 0.831     \\
MPRNet~\cite{Zamir2021MultiStagePI}                           & 24.69      & 0.707      & 0.347     & 8.34       & 0.365     & 0.798     \\
DGUNet~\cite{mou2022deep}                           & 19.67      & 0.627      & 0.384     & 8.88       & 0.391     & 0.810     \\
MIRV2~\cite{zamir2020learning}                         & 21.86      & 0.620      & 0.408     & 10.27      & 0.425     & 0.722     \\
SwinIR~\cite{liang2021swinir}                           & 17.72      & 0.661      & 0.419     & 6.89       & 0.301     & 0.852     \\
Restormer~\cite{zamir2022restormer}                        & 20.98      & 0.632      & 0.360     & 9.04       & 0.399     & 0.742     \\ \midrule
TAPE~\cite{liu2022tape}                             & 17.61      & 0.583      & 0.520     & 7.90       & 0.219     & 0.799     \\
AirNet~\cite{Li2022AllInOneIR}                         & 14.58      & 0.609      & 0.445     & 7.53       & 0.350     & 0.820     \\
DA-CLIP~\cite{luo2023controlling}                               & 15.74      & 0.606      & 0.472     & 14.91      & 0.475     & 0.739     \\
IDR~\cite{zhang2023ingredient}                              & 27.91      & 0.795      & 0.312     & 16.71      & 0.497     & 0.716     \\
DiffUIR~\cite{zheng2024selective}                          & 29.55      & 0.887      & 0.281     & 15.62      & 0.424     & {\color{red}0.505}     \\
\rowcolor[HTML]{E2E2E2} \textbf{VarFormer}                        & {\color{red}30.61}      & {\color{red}0.887}      & {\color{red}0.275}     & {\color{red}16.63}      & {\color{red}0.499}     & 0.605           \\ \bottomrule
\end{tabular}
\label{tab:zero-shot}
\end{table}

\subsection{Ablation Studies}
In this section, we perform a series of ablation studies to better the effectiveness of our designs. In Tab.~\ref{tab:alab_components}, we conduct ablation studies on the design of key components, including the adaptive mix-up skip that integrates encoder features into the decoder, the adapter that addresses the domain shift issue when using pre-trained VAR on degraded datasets, and the DAE and AFT modules responsible for generating and fusing multi-scale distribution alignment priors. We evaluate the performance based on the average performance across six tasks. The results indicate a gradual improvement with the addition of each component and a corresponding decline when each component is removed, underscoring the effectiveness of each module.

\begin{figure}[!ht]
\centering 
{\includegraphics[width=0.95\linewidth]{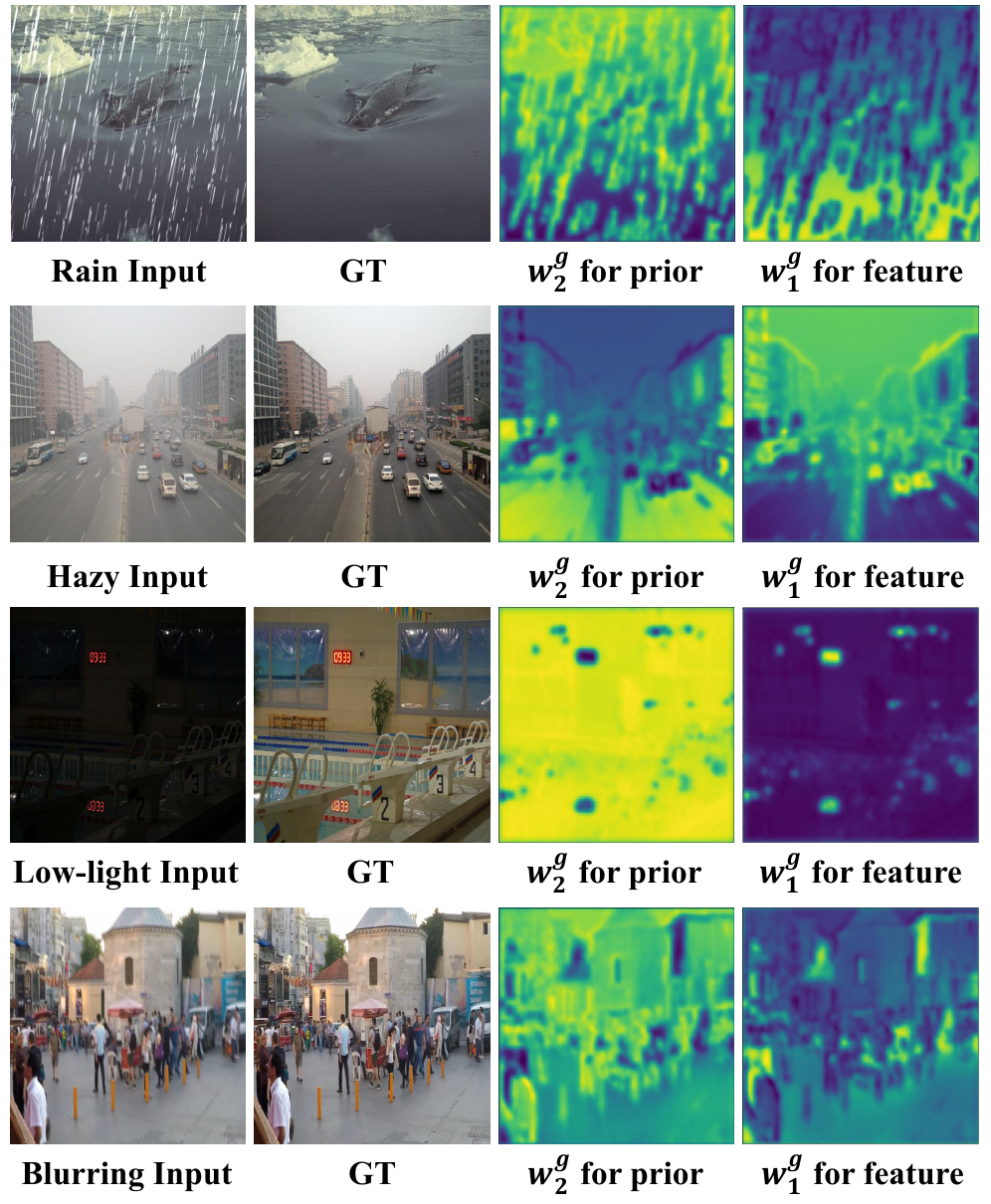}}
\caption{Visualization of weight maps from DAE.} 
\label{fig:f_visual}
\end{figure}

To verify the sensitivity of DAE module to various types of degradation, We visualize the region-specific fusion weights ${w}_{1}^{g}$ and ${w}_{2}^{g}$ within the DAE module in Fig.~\ref{fig:f_visual}. Here, weight ${w}_{1}^{g}$ is for the modulated feature map, and weight ${w}_{2}^{g}$ is for the VAR distribution alignment prior. It is observed that the areas highlighted by the VAR distribution alignment prior closely match the regions that have been adversely affected by degradation. This indicates that DAE module can accurately identify the degraded areas and effectively guide the introduction of high-quality distribution alignment priors for the restoration process.

\begin{table}[!h]
\centering
\caption{Ablation experiments on the components design. }
\setlength{\tabcolsep}{1.5mm}{
\begin{tabular}{c|cccc|cc}
\toprule
Exp. & skip       & adapter    & AFT        & DAE        & PSNR↑ & SSIM↑ \\ \midrule
a      &            &            &            &            & 27.54     & 0.841    \\
b      &            & \checkmark & \checkmark & \checkmark & 29.37     &  0.905     \\
c      & \checkmark &            & \checkmark & \checkmark &  29.51    & 0.917     \\
d      & \checkmark & \checkmark &            & \checkmark &  29.29    &  0.891    \\
e      & \checkmark &            & \checkmark &            &   28.98   &  0.867   \\
f      & \checkmark & \checkmark & \checkmark & \checkmark & 29.66     & 0.926     \\ \bottomrule

\end{tabular}}
\label{tab:alab_components}
\end{table}

\section{Conclusion}
\label{sec:conclusion}
In this paper, we investigate the multi-scale representations of the generative model VAR and reveal its endogenous multi-scale priors. As the autoregressive scales evolve, it transitions from capturing global color information to focusing on fine-grained details and adaptively aligns the input with clean images scale by scale. Furthermore, we propose the VarFormer framework integrated with multi-scale priors for multiple degradation restoration. Extensive experiments on various image restoration tasks demonstrate the effectiveness and generalization of our method.

\textbf{Acknowledgement.} This work was supported by the Anhui Provincial Natural Science Foundation under Grant 2108085UD12. We acknowledge the support of GPU cluster built by MCC Lab of Information Science and Technology Institution, USTC.

{
    \small
    \bibliographystyle{ieeenat_fullname}
    \bibliography{main}

\begin{thebibliography}{66}
\providecommand{\natexlab}[1]{#1}
\providecommand{\url}[1]{\texttt{#1}}
\expandafter\ifx\csname urlstyle\endcsname\relax
  \providecommand{\doi}[1]{doi: #1}\else
  \providecommand{\doi}{doi: \begingroup \urlstyle{rm}\Url}\fi

\bibitem[Abdal et~al.(2019)Abdal, Qin, and Wonka]{abdal2019image2stylegan}
Rameen Abdal, Yipeng Qin, and Peter Wonka.
\newblock Image2stylegan: How to embed images into the stylegan latent space?
\newblock In \emph{Proceedings of the IEEE/CVF international conference on computer vision}, pages 4432--4441, 2019.

\bibitem[Abdelhamed et~al.(2018)Abdelhamed, Lin, and Brown]{abdelhamed2018high}
Abdelrahman Abdelhamed, Stephen Lin, and Michael~S Brown.
\newblock A high-quality denoising dataset for smartphone cameras.
\newblock In \emph{Proceedings of the IEEE/CVF Conference on Computer Vision and Pattern Recognition}, pages 1692--1700, 2018.

\bibitem[Babacan et~al.(2009)Babacan, Molina, and Katsaggelos]{Babacan2009VariationalBB}
S.~Derin Babacan, Rafael Molina, and Aggelos~K. Katsaggelos.
\newblock Variational bayesian blind deconvolution using a total variation prior.
\newblock \emph{IEEE Transactions on Image Processing}, 18:\penalty0 12--26, 2009.

\bibitem[Brock(2018)]{brock2018large}
Andrew Brock.
\newblock Large scale gan training for high fidelity natural image synthesis.
\newblock \emph{arXiv preprint arXiv:1809.11096}, 2018.

\bibitem[Chen et~al.(2021)Chen, Lu, Zhang, Chu, and Chen]{chen2021hinet}
Liangyu Chen, Xin Lu, Jie Zhang, Xiaojie Chu, and Chengpeng Chen.
\newblock Hinet: Half instance normalization network for image restoration.
\newblock In \emph{Proceedings of the IEEE/CVF Conference on Computer Vision and Pattern Recognition}, pages 182--192, 2021.

\bibitem[Chen et~al.(2022)Chen, Chu, Zhang, and Sun]{chen2022simple}
Liangyu Chen, Xiaojie Chu, Xiangyu Zhang, and Jian Sun.
\newblock Simple baselines for image restoration.
\newblock In \emph{Proceedings of the European Conference on Computer Vision}, pages 17--33, 2022.

\bibitem[Chen et~al.(2024)Chen, Lu, Zhu, Ding, Yu, Ji, Li, Sun, Mao, and Zang]{chen2024sam2}
Tianrun Chen, Ankang Lu, Lanyun Zhu, Chaotao Ding, Chunan Yu, Deyi Ji, Zejian Li, Lingyun Sun, Papa Mao, and Ying Zang.
\newblock Sam2-adapter: Evaluating \& adapting segment anything 2 in downstream tasks: Camouflage, shadow, medical image segmentation, and more.
\newblock \emph{arXiv preprint arXiv:2408.04579}, 2024.

\bibitem[Feng et~al.(2018)Feng, Ji, Wang, Chang, Ren, and Gan]{feng2018challenges}
Weitao Feng, Deyi Ji, Yiru Wang, Shuorong Chang, Hansheng Ren, and Weihao Gan.
\newblock Challenges on large scale surveillance video analysis.
\newblock In \emph{Proceedings of the IEEE/CVF Conference on Computer Vision and Pattern Recognition Workshops}, pages 69--76, 2018.

\bibitem[Franzen(1999)]{franzen1999kodak}
Rich Franzen.
\newblock {Kodak lossless true color image suite}.
\newblock {\url{http://r0k.us/graphics/kodak}}, 1999.
\newblock {Accessed: 1999-06-07}.

\bibitem[Fu et~al.(2016)Fu, Huang, Ding, Liao, and Paisley]{Fu2016ClearingTS}
Xueyang Fu, Jiabin Huang, Xinghao Ding, Yinghao Liao, and John~William Paisley.
\newblock Clearing the skies: A deep network architecture for single-image rain removal.
\newblock \emph{IEEE Transactions on Image Processing}, 26:\penalty0 2944--2956, 2016.

\bibitem[Fu et~al.(2017)Fu, Huang, Zeng, Huang, Ding, and Paisley]{Fu2017RemovingRF}
Xueyang Fu, Jiabin Huang, Delu Zeng, Yue Huang, Xinghao Ding, and John~William Paisley.
\newblock Removing rain from single images via a deep detail network.
\newblock In \emph{Proceedings of the IEEE/CVF Conference on Computer Vision and Pattern Recognition}, pages 1715--1723, 2017.

\bibitem[Gu et~al.(2020)Gu, Shen, and Zhou]{gu2020image}
Jinjin Gu, Yujun Shen, and Bolei Zhou.
\newblock Image processing using multi-code gan prior.
\newblock In \emph{Proceedings of the IEEE/CVF Conference on Computer Vision and Pattern Recognition}, pages 3012--3021, 2020.

\bibitem[Guo et~al.(2020)Guo, Li, Guo, Loy, Hou, Kwong, and Cong]{guo2020zero}
Chunle Guo, Chongyi Li, Jichang Guo, Chen~Change Loy, Junhui Hou, Sam Kwong, and Runmin Cong.
\newblock Zero-reference deep curve estimation for low-light image enhancement.
\newblock In \emph{Proceedings of the IEEE/CVF Conference on Computer Vision and Pattern Recognition}, pages 1780--1789, 2020.

\bibitem[He et~al.(2010)He, Sun, and Tang]{he2010single}
Kaiming He, Jian Sun, and Xiaoou Tang.
\newblock Single image haze removal using dark channel prior.
\newblock \emph{IEEE Transactions on Pattern Analysis and Machine Intelligence}, 33\penalty0 (12):\penalty0 2341--2353, 2010.

\bibitem[Huang et~al.(2015)Huang, Singh, and Ahuja]{Huang2015SingleIS}
Jia-Bin Huang, Abhishek Singh, and Narendra Ahuja.
\newblock Single image super-resolution from transformed self-exemplars.
\newblock In \emph{Proceedings of the IEEE/CVF Conference on Computer Vision and Pattern Recognition}, pages 5197--5206, 2015.

\bibitem[Ji et~al.(2023)Ji, Zhao, Lu, Tao, and Ye]{urur}
Deyi Ji, Feng Zhao, Hongtao Lu, Mingyuan Tao, and Jieping Ye.
\newblock Ultra-high resolution segmentation with ultra-rich context: A novel benchmark.
\newblock In \emph{Proceedings of the IEEE/CVF Conference on Computer Vision and Pattern Recognition}, pages 23621--23630, 2023.

\bibitem[Karras(2017)]{karras2017progressive}
Tero Karras.
\newblock Progressive growing of gans for improved quality, stability, and variation.
\newblock \emph{arXiv preprint arXiv:1710.10196}, 2017.

\bibitem[Karras et~al.(2019)Karras, Laine, and Aila]{karras2019style}
Tero Karras, Samuli Laine, and Timo Aila.
\newblock A style-based generator architecture for generative adversarial networks.
\newblock In \emph{Proceedings of the IEEE/CVF Conference on Computer Vision and Pattern Recognition}, pages 4401--4410, 2019.

\bibitem[Karras et~al.(2020)Karras, Laine, Aittala, Hellsten, Lehtinen, and Aila]{karras2020analyzing}
Tero Karras, Samuli Laine, Miika Aittala, Janne Hellsten, Jaakko Lehtinen, and Timo Aila.
\newblock Analyzing and improving the image quality of stylegan.
\newblock In \emph{Proceedings of the IEEE/CVF Conference on Computer Vision and Pattern Recognition}, pages 8110--8119, 2020.

\bibitem[Kawar et~al.(2022)Kawar, Elad, Ermon, and Song]{kawar2022denoising}
Bahjat Kawar, Michael Elad, Stefano Ermon, and Jiaming Song.
\newblock Denoising diffusion restoration models.
\newblock \emph{Advances in Neural Information Processing Systems}, 35:\penalty0 23593--23606, 2022.

\bibitem[Kingma and Ba(2014)]{Kingma2014AdamAM}
Diederik~P. Kingma and Jimmy Ba.
\newblock Adam: A method for stochastic optimization.
\newblock \emph{arXiv preprint arXiv:1412.6980}, 2014.

\bibitem[Li et~al.(2018{\natexlab{a}})Li, Ren, Fu, Tao, Feng, Zeng, and Wang]{li2018benchmarking}
Boyi Li, Wenqi Ren, Dengpan Fu, Dacheng Tao, Dan Feng, Wenjun Zeng, and Zhangyang Wang.
\newblock Benchmarking single-image dehazing and beyond.
\newblock \emph{IEEE Transactions on Image Processing}, 28\penalty0 (1):\penalty0 492--505, 2018{\natexlab{a}}.

\bibitem[Li et~al.(2022)Li, Liu, Hu, Wu, Lv, and Peng]{Li2022AllInOneIR}
Boyun Li, Xiao Liu, Peng Hu, Zhongqin Wu, Jiancheng Lv, and Xiaocui Peng.
\newblock All-in-one image restoration for unknown corruption.
\newblock In \emph{Proceedings of the IEEE/CVF Conference on Computer Vision and Pattern Recognition}, pages 17431--17441, 2022.

\bibitem[Li et~al.(2018{\natexlab{b}})Li, Wu, Lin, Liu, and Zha]{Li2018RecurrentSC}
Xia Li, Jianlong Wu, Zhouchen Lin, Hong Liu, and Hongbin Zha.
\newblock Recurrent squeeze-and-excitation context aggregation net for single image deraining.
\newblock \emph{arXiv preprint arXiv:1807.05698}, 2018{\natexlab{b}}.

\bibitem[Li et~al.(2016)Li, Tan, Guo, Lu, and Brown]{Li2016RainSR}
Yu Li, Robby~T. Tan, Xiaojie Guo, Jiangbo Lu, and M.~S. Brown.
\newblock Rain streak removal using layer priors.
\newblock In \emph{Proceedings of the IEEE/CVF Conference on Computer Vision and Pattern Recognition}, pages 2736--2744, 2016.

\bibitem[Liang et~al.(2021)Liang, Cao, Sun, Zhang, Van~Gool, and Timofte]{liang2021swinir}
Jingyun Liang, Jiezhang Cao, Guolei Sun, Kai Zhang, Luc Van~Gool, and Radu Timofte.
\newblock Swinir: Image restoration using swin transformer.
\newblock In \emph{Proceedings of the IEEE/CVF International Conference on Computer Vision}, pages 1833--1844, 2021.

\bibitem[Lin et~al.(2023)Lin, He, Chen, Lyu, Dai, Yu, Ouyang, Qiao, and Dong]{lin2023diffbir}
Xinqi Lin, Jingwen He, Ziyan Chen, Zhaoyang Lyu, Bo Dai, Fanghua Yu, Wanli Ouyang, Yu Qiao, and Chao Dong.
\newblock Diffbir: Towards blind image restoration with generative diffusion prior.
\newblock \emph{arXiv preprint arXiv:2308.15070}, 2023.

\bibitem[Liu et~al.(2024{\natexlab{a}})Liu, Wang, Fan, Wang, Tang, and Qu]{liu2024residual}
Jiawei Liu, Qiang Wang, Huijie Fan, Yinong Wang, Yandong Tang, and Liangqiong Qu.
\newblock Residual denoising diffusion models.
\newblock In \emph{Proceedings of the IEEE/CVF Conference on Computer Vision and Pattern Recognition}, pages 2773--2783, 2024{\natexlab{a}}.

\bibitem[Liu et~al.(2022)Liu, Xie, Zhang, Yuan, Chen, Zhou, Li, and Tian]{liu2022tape}
Lin Liu, Lingxi Xie, Xiaopeng Zhang, Shanxin Yuan, Xiangyu Chen, Wengang Zhou, Houqiang Li, and Qi Tian.
\newblock Tape: Task-agnostic prior embedding for image restoration.
\newblock In \emph{Proceedings of the European Conference on Computer Vision}, pages 447--464, 2022.

\bibitem[Liu et~al.(2024{\natexlab{b}})Liu, Li, Ma, Xie, and Liu]{liu2024hcanet}
Yi Liu, Jiachen Li, Yanchun Ma, Qing Xie, and Yongjian Liu.
\newblock Hcanet: Haze-concentration-aware network for real-scene dehazing with codebook priors.
\newblock In \emph{Proceedings of the 32nd ACM International Conference on Multimedia}, pages 9136--9144, 2024{\natexlab{b}}.

\bibitem[Luo et~al.(2015)Luo, Xu, and Ji]{Luo2015RemovingRF}
Yu Luo, Yong Xu, and Hui Ji.
\newblock Removing rain from a single image via discriminative sparse coding.
\newblock In \emph{Proceedings of the IEEE/CVF International Conference on Computer Vision}, pages 3397--3405, 2015.

\bibitem[Luo et~al.(2023)Luo, Gustafsson, Zhao, Sj{\"o}lund, and Sch{\"o}n]{luo2023controlling}
Ziwei Luo, Fredrik~K Gustafsson, Zheng Zhao, Jens Sj{\"o}lund, and Thomas~B Sch{\"o}n.
\newblock Controlling vision-language models for universal image restoration.
\newblock \emph{arXiv preprint arXiv:2310.01018}, 2023.

\bibitem[Ma et~al.(2016)Ma, Duanmu, Wu, Wang, Yong, Li, and Zhang]{ma2016waterloo}
Kede Ma, Zhengfang Duanmu, Qingbo Wu, Zhou Wang, Hongwei Yong, Hongliang Li, and Lei Zhang.
\newblock Waterloo exploration database: New challenges for image quality assessment models.
\newblock \emph{IEEE Transactions on Image Processing}, 26\penalty0 (2):\penalty0 1004--1016, 2016.

\bibitem[Martin et~al.(2001)Martin, Fowlkes, Tal, and Malik]{martin2001database}
David Martin, Charless Fowlkes, Doron Tal, and Jitendra Malik.
\newblock A database of human segmented natural images and its application to evaluating segmentation algorithms and measuring ecological statistics.
\newblock In \emph{Proceedings of the IEEE/CVF International Conference on Computer Vision}, pages 416--423, 2001.

\bibitem[Mou et~al.(2022)Mou, Wang, and Zhang]{mou2022deep}
Chong Mou, Qian Wang, and Jian Zhang.
\newblock Deep generalized unfolding networks for image restoration.
\newblock In \emph{Proceedings of the IEEE/CVF Conference on Computer Vision and Pattern Recognition}, pages 17399--17410, 2022.

\bibitem[Nah et~al.(2016)Nah, Kim, and Lee]{Nah2016DeepMC}
Seungjun Nah, Tae~Hyun Kim, and Kyoung~Mu Lee.
\newblock Deep multi-scale convolutional neural network for dynamic scene deblurring.
\newblock In \emph{Proceedings of the IEEE/CVF Conference on Computer Vision and Pattern Recognition}, pages 257--265, 2016.

\bibitem[Potlapalli et~al.(2023)Potlapalli, Zamir, Khan, and Khan]{potlapalli2306promptir}
V Potlapalli, SW Zamir, S Khan, and FS Khan.
\newblock Promptir: Prompting for all-in-one blind image restoration.
\newblock \emph{arXiv preprint arXiv:2306.13090}, 2023.

\bibitem[Quan et~al.(2023)Quan, Wu, and Ji]{quan2023neumann}
Yuhui Quan, Zicong Wu, and Hui Ji.
\newblock Neumann network with recursive kernels for single image defocus deblurring.
\newblock In \emph{Proceedings of the IEEE/CVF Conference on Computer Vision and Pattern Recognition}, pages 5754--5763, 2023.

\bibitem[Ruan et~al.(2022)Ruan, Chen, Li, and Lam]{ruan2022learning}
Lingyan Ruan, Bin Chen, Jizhou Li, and Miuling Lam.
\newblock Learning to deblur using light field generated and real defocus images.
\newblock In \emph{Proceedings of the IEEE/CVF Conference on Computer Vision and Pattern Recognition}, pages 16304--16313, 2022.

\bibitem[Tian et~al.(2024)Tian, Jiang, Yuan, Peng, and Wang]{tian2024visual}
Keyu Tian, Yi Jiang, Zehuan Yuan, Bingyue Peng, and Liwei Wang.
\newblock Visual autoregressive modeling: Scalable image generation via next-scale prediction.
\newblock \emph{arXiv preprint arXiv:2404.02905}, 2024.

\bibitem[Tsai et~al.(2022)Tsai, Peng, Lin, Tsai, and Lin]{tsai2022stripformer}
Fu-Jen Tsai, Yan-Tsung Peng, Yen-Yu Lin, Chung-Chi Tsai, and Chia-Wen Lin.
\newblock Stripformer: Strip transformer for fast image deblurring.
\newblock In \emph{Proceedings of the European Conference on Computer Vision}, pages 146--162, 2022.

\bibitem[Tu et~al.(2022)Tu, Talebi, Zhang, Yang, Milanfar, Bovik, and Li]{tu2022maxim}
Zhengzhong Tu, Hossein Talebi, Han Zhang, Feng Yang, Peyman Milanfar, Alan Bovik, and Yinxiao Li.
\newblock Maxim: Multi-axis mlp for image processing.
\newblock In \emph{Proceedings of the IEEE/CVF Conference on Computer Vision and Pattern Recognition}, pages 5769--5780, 2022.

\bibitem[Valanarasu et~al.(2022)Valanarasu, Yasarla, and Patel]{valanarasu2022transweather}
Jeya Maria~Jose Valanarasu, Rajeev Yasarla, and Vishal~M Patel.
\newblock Transweather: Transformer-based restoration of images degraded by adverse weather conditions.
\newblock In \emph{Proceedings of the IEEE/CVF Conference on Computer Vision and Pattern Recognition}, pages 2353--2363, 2022.

\bibitem[Wang et~al.(2023{\natexlab{a}})Wang, Yue, Zhou, Chan, and Loy]{wang2024exploiting}
Jianyi Wang, Zongsheng Yue, Shangchen Zhou, Kelvin~CK Chan, and Chen~Change Loy.
\newblock Exploiting diffusion prior for real-world image super-resolution.
\newblock \emph{arXiv preprint arXiv:2305.07015}, 2023{\natexlab{a}}.

\bibitem[Wang et~al.(2023{\natexlab{b}})Wang, Wang, Cao, Shen, and Huang]{wang2023images}
Xinlong Wang, Wen Wang, Yue Cao, Chunhua Shen, and Tiejun Huang.
\newblock Images speak in images: A generalist painter for in-context visual learning.
\newblock In \emph{Proceedings of the IEEE/CVF Conference on Computer Vision and Pattern Recognition}, pages 6830--6839, 2023{\natexlab{b}}.

\bibitem[Wang et~al.(2022)Wang, Yu, and Zhang]{wang2022zero}
Yinhuai Wang, Jiwen Yu, and Jian Zhang.
\newblock Zero-shot image restoration using denoising diffusion null-space model.
\newblock \emph{arXiv preprint arXiv:2212.00490}, 2022.

\bibitem[Wang et~al.(2021)Wang, Cun, Bao, and Liu]{Wang2021UformerAG}
Zhendong Wang, Xiaodong Cun, Jianmin Bao, and Jianzhuang Liu.
\newblock Uformer: A general u-shaped transformer for image restoration.
\newblock In \emph{Proceedings of the IEEE/CVF Conference on Computer Vision and Pattern Recognition}, pages 17662--17672, 2021.

\bibitem[Wang et~al.(2023{\natexlab{c}})Wang, Fu, Liu, and Zhang]{wang2023lg}
Zichun Wang, Ying Fu, Ji Liu, and Yulun Zhang.
\newblock Lg-bpn: Local and global blind-patch network for self-supervised real-world denoising.
\newblock In \emph{Proceedings of the IEEE/CVF Conference on Computer Vision and Pattern Recognition}, pages 18156--18165, 2023{\natexlab{c}}.

\bibitem[Wei et~al.(2018)Wei, Wang, Yang, and Liu]{Wei2018DeepRD}
Chen Wei, Wenjing Wang, Wenhan Yang, and Jiaying Liu.
\newblock Deep retinex decomposition for low-light enhancement.
\newblock \emph{arXiv preprint arXiv:1808.04560}, 2018.

\bibitem[Wu et~al.(2023)Wu, Pan, Wang, Yang, Wei, Li, and Shen]{wu2023learning}
Yuhui Wu, Chen Pan, Guoqing Wang, Yang Yang, Jiwei Wei, Chongyi Li, and Heng~Tao Shen.
\newblock Learning semantic-aware knowledge guidance for low-light image enhancement.
\newblock In \emph{Proceedings of the IEEE/CVF Conference on Computer Vision and Pattern Recognition}, pages 1662--1671, 2023.

\bibitem[Xu et~al.(2023)Xu, Wang, and Lu]{xu2023low}
Xiaogang Xu, Ruixing Wang, and Jiangbo Lu.
\newblock Low-light image enhancement via structure modeling and guidance.
\newblock In \emph{Proceedings of the IEEE/CVF Conference on Computer Vision and Pattern Recognition}, pages 9893--9903, 2023.

\bibitem[Yang et~al.(2016)Yang, Tan, Feng, Liu, Guo, and Yan]{Yang2016DeepJR}
Wenhan Yang, Robby~T. Tan, Jiashi Feng, Jiaying Liu, Zongming Guo, and Shuicheng Yan.
\newblock Deep joint rain detection and removal from a single image.
\newblock In \emph{Proceedings of the IEEE/CVF Conference on Computer Vision and Pattern Recognition}, pages 1685--1694, 2016.

\bibitem[Yang et~al.(2019)Yang, Tan, Wang, Fang, and Liu]{Yang2019SingleID}
Wenhan Yang, Robby~T. Tan, Shiqi Wang, Yuming Fang, and Jiaying Liu.
\newblock Single image deraining: From model-based to data-driven and beyond.
\newblock \emph{IEEE Transactions on Pattern Analysis and Machine Intelligence}, 43:\penalty0 4059--4077, 2019.

\bibitem[Ye et~al.(2023)Ye, Chen, Bai, Shi, Xue, Jiang, Yin, Chen, and Liu]{ye2023adverse}
Tian Ye, Sixiang Chen, Jinbin Bai, Jun Shi, Chenghao Xue, Jingxia Jiang, Junjie Yin, Erkang Chen, and Yun Liu.
\newblock Adverse weather removal with codebook priors.
\newblock In \emph{Proceedings of the IEEE/CVF International Conference on Computer Vision}, pages 12653--12664, 2023.

\bibitem[Zamir et~al.(2020)Zamir, Arora, Khan, Hayat, Khan, Yang, and Shao]{zamir2020learning}
Syed~Waqas Zamir, Aditya Arora, Salman Khan, Munawar Hayat, Fahad~Shahbaz Khan, Ming-Hsuan Yang, and Ling Shao.
\newblock Learning enriched features for real image restoration and enhancement.
\newblock In \emph{Proceedings of the European Conference on Computer Vision}, pages 492--511, 2020.

\bibitem[Zamir et~al.(2021)Zamir, Arora, Khan, Hayat, Khan, Yang, and Shao]{Zamir2021MultiStagePI}
Syed~Waqas Zamir, Aditya Arora, Salman~Hameed Khan, Munawar Hayat, Fahad~Shahbaz Khan, Ming-Hsuan Yang, and Ling Shao.
\newblock Multi-stage progressive image restoration.
\newblock In \emph{Proceedings of the IEEE/CVF Conference on Computer Vision and Pattern Recognition}, pages 14816--14826, 2021.

\bibitem[Zamir et~al.(2022)Zamir, Arora, Khan, Hayat, Khan, and Yang]{zamir2022restormer}
Syed~Waqas Zamir, Aditya Arora, Salman Khan, Munawar Hayat, Fahad~Shahbaz Khan, and Ming-Hsuan Yang.
\newblock Restormer: Efficient transformer for high-resolution image restoration.
\newblock In \emph{Proceedings of the IEEE/CVF Conference on Computer Vision and Pattern Recognition}, pages 5728--5739, 2022.

\bibitem[Zhang and Patel(2018)]{Zhang2018DensityAwareSI}
He Zhang and Vishal~M. Patel.
\newblock Density-aware single image de-raining using a multi-stream dense network.
\newblock In \emph{Proceedings of the IEEE/CVF Conference on Computer Vision and Pattern Recognition}, pages 695--704, 2018.

\bibitem[Zhang et~al.(2017{\natexlab{a}})Zhang, Sindagi, and Patel]{Zhang2017ImageDU}
He Zhang, Vishwanath~A. Sindagi, and Vishal~M. Patel.
\newblock Image de-raining using a conditional generative adversarial network.
\newblock \emph{IEEE Transactions on Circuits and Systems for Video Technology}, 30:\penalty0 3943--3956, 2017{\natexlab{a}}.

\bibitem[Zhang et~al.(2022{\natexlab{a}})Zhang, Zhang, Gu, Zhang, Kong, and Yuan]{zhang2022accurate}
Jiale Zhang, Yulun Zhang, Jinjin Gu, Yongbing Zhang, Linghe Kong, and Xin Yuan.
\newblock Accurate image restoration with attention retractable transformer.
\newblock \emph{arXiv preprint arXiv:2210.01427}, 2022{\natexlab{a}}.

\bibitem[Zhang et~al.(2023)Zhang, Huang, Yao, Yang, Yu, Zhou, and Zhao]{zhang2023ingredient}
Jinghao Zhang, Jie Huang, Mingde Yao, Zizheng Yang, Hu Yu, Man Zhou, and Feng Zhao.
\newblock Ingredient-oriented multi-degradation learning for image restoration.
\newblock In \emph{Proceedings of the IEEE/CVF Conference on Computer Vision and Pattern Recognition}, pages 5825--5835, 2023.

\bibitem[Zhang et~al.(2017{\natexlab{b}})Zhang, Zuo, Chen, Meng, and Zhang]{zhang2017beyond}
Kai Zhang, Wangmeng Zuo, Yunjin Chen, Deyu Meng, and Lei Zhang.
\newblock Beyond a gaussian denoiser: Residual learning of deep cnn for image denoising.
\newblock \emph{IEEE Transactions on Image Processing}, 26\penalty0 (7):\penalty0 3142--3155, 2017{\natexlab{b}}.

\bibitem[Zhang et~al.(2022{\natexlab{b}})Zhang, Li, Liang, Cao, Zhang, Tang, Timofte, and Van~Gool]{zhang2022practical}
Kai Zhang, Yawei Li, Jingyun Liang, Jiezhang Cao, Yulun Zhang, Hao Tang, Radu Timofte, and Luc Van~Gool.
\newblock Practical blind denoising via swin-conv-unet and data synthesis.
\newblock \emph{arXiv preprint arXiv:2203.13278}, 2022{\natexlab{b}}.

\bibitem[Zhang et~al.(2018)Zhang, Isola, Efros, Shechtman, and Wang]{zhang2018unreasonable}
Richard Zhang, Phillip Isola, Alexei~A Efros, Eli Shechtman, and Oliver Wang.
\newblock The unreasonable effectiveness of deep features as a perceptual metric.
\newblock In \emph{Proceedings of the IEEE/CVF Conference on Computer Vision and Pattern Recognition}, pages 586--595, 2018.

\bibitem[Zheng et~al.(2024)Zheng, Wu, Yang, Zhang, Hu, and Zheng]{zheng2024selective}
Dian Zheng, Xiao-Ming Wu, Shuzhou Yang, Jian Zhang, Jian-Fang Hu, and Wei-Shi Zheng.
\newblock Selective hourglass mapping for universal image restoration based on diffusion model.
\newblock In \emph{Proceedings of the IEEE/CVF Conference on Computer Vision and Pattern Recognition}, pages 25445--25455, 2024.

\bibitem[Zhou et~al.(2021)Zhou, Ren, Emerton, Lim, and Large]{zhou2021image}
Yuqian Zhou, David Ren, Neil Emerton, Sehoon Lim, and Timothy Large.
\newblock Image restoration for under-display camera.
\newblock In \emph{Proceedings of the IEEE/CVF Conference on Computer Vision and Pattern Recognition}, pages 9179--9188, 2021.

\end{thebibliography}

}

\end{document}